%% file: conference.tex
\documentclass[5p]{elsarticle}

\usepackage{amsmath,amssymb,amsfonts}
\usepackage{algorithmic}
\usepackage{graphicx}
\usepackage{textcomp}
\usepackage{tikz}
\usepackage{xcolor}
\usepackage{url}
\usepackage{hyperref}
\begin{document}
	
	\title{Fruit Fly Classification (Diptera: Tephritidae) in Images, Applying Transfer Learning}
	
	\author[unsaac]{Erick Andrew Bustamante Flores\corref{cor1}}
	\ead{eerick28@gmail.com}
	
	\author[unsaac]{Harley Vera-Olivera}
	\ead{harley.vera@unsaac.edu.pe}
	
	\author[unsaac]{Ivan Cesar Medrano Valencia}
	\ead{ivan.medrano@unsaac.edu.pe}
	
	\author[unsaac]{Carlos Fernando Montoya Cubas}
	\ead{carlos.montoya@unsaac.edu.pe}
	
	\cortext[cor1]{Corresponding author}
	
	\address[unsaac]{Department of Computer Science, Universidad Nacional de San Antonio Abad del Cusco, Cusco, Perú}

	\input{contenido/abstract}

	\maketitle
	
	\input{contenido/Introduccion}
	\input{contenido/antecedentes/main_antecedentes}
	\input{contenido/mosca_fruta/main_mosca_fruta}
	\input{contenido/materiales_metodos/main_mat_met}

	\input{contenido/experimentos_resultados/main_exp_res}
	\input{contenido/discusion/main_discus}
	\input{contenido/conclusiones/main_conclusiones}
	\input{contenido/agradecimientos/agradecimientos}

	\bibliography{referencias}
	
\end{document}

%% file: contenido/abstract.tex
\begin{abstract}
This study develops a transfer learning model for the automated classification of two species of fruit flies, \textit{Anastrepha fraterculus} and \textit{Ceratitis capitata}, in a controlled laboratory environment. The research addresses the need to optimize identification and classification, which are currently performed manually by experts, being affected by human factors and facing time challenges. The methodological process of this study includes the capture of high-quality images using a mobile phone camera and a stereo microscope, followed by segmentation to reduce size and focus on relevant morphological areas. The images were carefully labeled and preprocessed to ensure the quality and consistency of the dataset used to train the pre-trained convolutional neural network models VGG16, VGG19, and Inception-v3. The results were evaluated using the F1-score, achieving 82\% for VGG16 and VGG19, while Inception-v3 reached an F1-score of 93\%. Inception-v3's reliability was verified through model testing in uncontrolled environments, with positive results, complemented by the Grad-CAM technique, demonstrating its ability to capture essential morphological features. These findings indicate that Inception-v3 is an effective and replicable approach for classifying \textit{A. fraterculus} and \textit{C. capitata}, with potential for implementation in automated monitoring systems.
\end{abstract}


%% file: contenido/Introduccion.tex
\section{Introduction}
Fruit flies, belonging to the Order Diptera and the Family Tephritidae, are one of the main pests affecting fruit production worldwide. There are approximately 4,000 species of fruit flies, distributed in genera of great economic importance such as \textit{Anastrepha}, \textit{Bactrocera}, \textit{Ceratitis}, \textit{Dacus}, \textit{Rhagoletis}, and \textit{Toxotrypana} \cite{3ortiz2021, delgado2009moscas}. In Peru, several genera are present, including \textit{Anastrepha} (with 44 species), \textit{Ceratitis} (1 species), \textit{Rhagoletis} (5 species), and \textit{Toxotrypana} (1 species) \cite{iica1997}. The species of greatest economic relevance, such as \textit{Anastrepha distincta}, \textit{Anastrepha fraterculus}, \textit{Ceratitis capitata}, and \textit{Anastrepha striata}, are found in various regions of the country, from the south and central areas to the north \cite{senasa2011}.

The biological cycle of these pests causes significant damage to fruit production. Females deposit their eggs under the skin of the fruit, where the larvae develop and consume the pulp, accelerating ripening, decomposition, and falling, leading to substantial economic losses. In Latin America, fruit flies are estimated to cause annual losses of around 35 million dollars, and in Andean Group countries (Bolivia, Colombia, Ecuador, and Peru), these losses exceed 30\% of the value of fruit and vegetable production \cite{1municipalidad_echarati}. In Peru, the damage caused by these pests reaches up to 60\% of the production \cite{2senasa_erradicacion}.

To mitigate these losses, ``El Servicio Nacional de Sanidad Agraria del Perú'' (SENASA) has implemented fruit fly detection and monitoring projects. This system relies on the use of McPhail traps with food attractants, which allow for the collection of specimens for laboratory analysis. Specifically, in the province of ``La Convención'' in the Cusco region, specialists identify and classify species by analyzing specific morphological characteristics, and weekly reports are generated to evaluate the population of these species in the monitored areas.

Currently, species identification is performed manually by experts, which requires speed and precision. However, both factors are compromised: the average identification time per fly is around 10 seconds, and the accuracy of the analysis can be affected by human factors such as fatigue and subjectivity. This manual process, although necessary, is not optimal during peak fly population periods, resulting in delays in decision-making.

Recent studies have shown that computational systems based on artificial intelligence (AI), particularly those using deep learning techniques, are capable of improving accuracy in similar classification processes \cite{4Das2018, 22Toke2020}. The literature suggests that by using datasets combined with machine learning and deep learning models, accuracy rates exceeding 85\% can be achieved \cite{7Faria2014, 9Martins2019, 13Leonardo2017}. However, to date, these methods have not been specifically applied to \textit{Anastrepha fraterculus} and \textit{Ceratitis capitata} species, nor has the potential of a dataset based on images taken with a mobile phone through a stereomicroscope been explored.

This study proposes the development of a computational model for the identification and classification of \textit{Anastrepha fraterculus} and \textit{Ceratitis capitata} through image analysis. The main contributions include the creation of an original dataset and the development of an automated classification model. The dataset was obtained from samples selected by an expert, using a stereomicroscope and a mobile device. The proposed model is based on the analysis of key morphological areas of each species, replicating the identification process followed by specialists in the laboratory. This methodology has the potential to be implemented in AI-assisted taxonomic classification environments, thereby optimizing monitoring and control systems for this pest.

The organization of this paper is as follows: Section 2 reviews related works; Section 3 describes the morphological characteristics of the studied species; Section 4 presents the materials and methods; Section 5 details the experiments and results obtained; Section 6 discusses the findings; and finally, Section 7 presents the conclusions and analyzes the study's findings.

%% file: contenido/antecedentes/main_antecedentes.tex
\section{Antecedentes}
\input{contenido/antecedentes/ML_clasificacion}
\input{contenido/antecedentes/DL_clasificacion}
\input{contenido/antecedentes/DL_ML_clasificacion}

%% file: contenido/antecedentes/ML_clasificacion.tex
\subsection{Machine Learning Techniques for Fruit Fly Classification}

In the study by \cite{7Faria2014}, the classification of the fruit fly species \textit{Anastrepha fraterculus}, \textit{Anastrepha obliqua}, and \textit{Anastrepha sororcula} was investigated using a set of wing and ovipositor images. During preprocessing, Otsu's thresholding and morphological dilation were employed to segment the images, extracting color and texture features to train multiple classifiers. Accuracy was optimized by combining up to 33 individual classifiers with a Support Vector Machines (SVM) meta-classifier and merging the datasets, achieving a final accuracy of 98.8\%.

\cite{8Remboski2018} proposed the development of a classification system designed to identify fruit fly species using smart traps, utilizing two datasets: one with four classes (\textit{Anastrepha fraterculus}, \textit{Ceratitis capitata}, "Other insects," and "Residues"), and another with three classes, consolidating the last two into "Others." The images were converted to grayscale, adaptive thresholding and morphological opening were applied to remove unwanted areas. They used Bag of Visual Words (BOVW) for feature extraction and evaluated various classification models, with SVM being the most effective, achieving accuracies of 84.56\% and 86.38\% in each dataset.

%% file: contenido/antecedentes/DL_clasificacion.tex
\subsection{Deep Learning Techniques for Fruit Fly Classification}

The classification of \textit{Ceratitis capitata} and \textit{Grapholita molesta} in smart traps was addressed by \cite{9Martins2019} using pre-trained models, with ResNet18 standing out, initially achieving an accuracy of 84.28\%. The accuracy improved to 93.55\% and 91.28\%, respectively, after applying data augmentation techniques (flipping, rotation, and random erasing). Given the goal of implementing a model on the Raspberry Pi v2, SqueezeNet was selected, achieving an accuracy of 88.56\% for \textit{Ceratitis capitata} and 90.60\% for \textit{Grapholita molesta} with the same data augmentation techniques.

\cite{13Leonardo2017} focused on the identification of fruit fly species, using a dataset containing wing images of \textit{Anastrepha fraterculus}, \textit{Anastrepha obliqua}, and \textit{Anastrepha sororcula}.

Feature descriptors and detectors were employed to train nine machine learning techniques. The best performance was achieved with the Multilayer Perceptron (MLP), which reached an average accuracy of 88.9\% in classifying all species.

The research by \cite{15Gonzalez-Lopez2022} addressed the determination of the age of fruit fly pupae with the goal of making them sterile. The dataset was prepared by cleaning a section of the pupae to expose the fly’s eyes.

The Multi-Template Matching technique was applied alongside an Inception-v1 model. This combination of image preprocessing with templates and the Inception-v1 model achieved an accuracy of 72.22\% for \textit{Anastrepha ludens} and 83.17\% for \textit{Ceratitis capitata}.

%% file: contenido/antecedentes/DL_ML_clasificacion.tex
\subsection{Deep Learning Techniques Combined with Machine Learning for Fruit Fly Classification}

The research by \cite{10Peng2020} addressed the classification of \textit{Bactrocera dorsalis}, \textit{Bactrocera cucurbitae}, \textit{Bactrocera tau}, and \textit{Bactrocera scutellata} in images with complex backgrounds. Initially, they tested different hyperparameter configurations in a CNN but faced overfitting issues. To resolve this, they replaced the final Softmax layer of the CNN with machine learning models: SVM, KNN, AdaBoost, and Random Forest. The CNN-SVM combination was the most effective, achieving an accuracy of 92.4\%.

In their work, \cite{12Leonardo2018} successfully classified \textit{Anastrepha} species: \textit{A. fraterculus}, \textit{A. obliqua}, and \textit{A. sororcula}, using pre-trained models (ResNet, VGG16, VGG19, Xception, Inception) to extract features. These features were evaluated with machine learning models such as Decision Tree, KNN, MLP, Naive Bayes, SGD, and SVM. Using cross-validation, VGG16-SVM achieved the highest accuracy, reaching 95.68\%.

The proposal by \cite{14Molina-Rotger2023} focused on the development of a monitoring system using sticky traps, with two datasets: one distinguishing "Olive Fly" and "Others" (including debris), and another differentiating "Fruit Fly" and debris. They used models such as Random Forest, SVM, Decision Trees, and deep networks like VGG16, MobileNet, and Xception. For the first dataset, RF and SVM achieved accuracies of 62.1\% and 86.4\%. In the second dataset, accuracies increased to 91.9\% (RF) and 94.5\% (SVM).

%% file: contenido/mosca_fruta/main_mosca_fruta.tex
\section{Morphological Characteristics.}
\label{sec:moscafruta}
Understanding the morphological characteristics of different species of fruit flies is crucial for appreciating the work of experts in the identification and classification of these pests in the laboratory. The accuracy in species identification depends on a thorough analysis of their morphological traits, among which the wings, thorax, and ovipositor are key elements. These characteristics allow differentiation of species that, although visually similar, exhibit subtle variations in their anatomical structures, essential for their taxonomic classification.

\subsection{Sections and Venations of the Fruit Fly Wing.}
The wings of the fruit fly provide key information through their sections and venations, which are crucial for the classification of different species. This information is represented in specific patterns and structural differences that allow effective identification and differentiation between species.

\begin{figure}[htbp]
    \centering
    \includegraphics[width=0.5\textwidth]{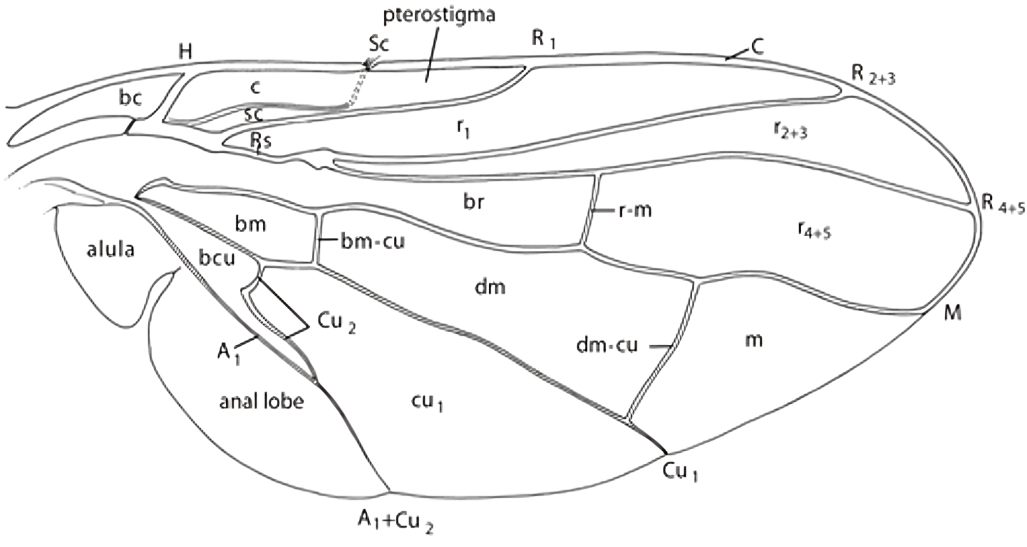}
    \caption{Sections of the wing.}
    \scriptsize \textit{Adapted from:} \cite{norrbom}.
    \label{fig:taxonomiaala}
\end{figure}

In Figure \ref{fig:taxonomiaala}, the distribution of the venations and sections present in the wing is shown. Some venations provide relevant information for species differentiation, such as the $R_{4+5}$, $bm{-}cu$, $dm{-}cu$ veins, among others. Similarly, specific sections such as $dm$, $r_{4+5}$, $bm$, among others, are identified. For the two species studied in this research, the venations and sections that allow their differentiation will be described in detail.

\subsection{Parts of the Fruit Fly Thorax.}
The thorax of the fruit fly constitutes another significant source of information due to its structure and the coloration patterns it exhibits. These characteristics are essential for distinguishing between various species, enabling precise classification.

\begin{figure}[htbp]
    \centering
    \includegraphics[width=0.4\textwidth]{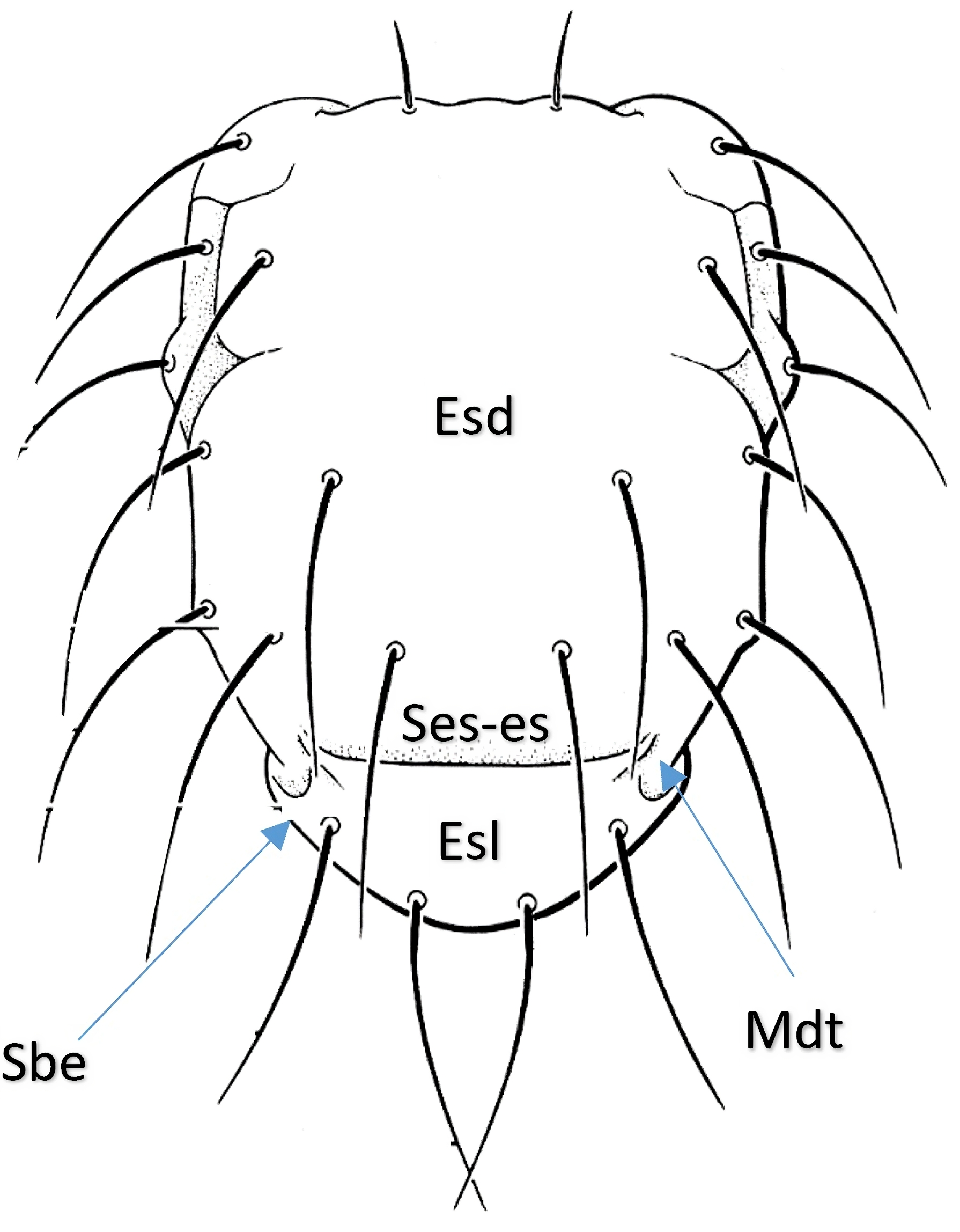}
    \caption{Parts of the Thorax}
    \scriptsize \textit{Adapted from:} \cite{senasaTephritidae, especiesImEco}.
    \label{fig:taxonomiatorax}
\end{figure}

In Figure \ref{fig:taxonomiatorax}, the distribution of the parts that make up the thorax is presented. Some anatomical structures provide key information for species differentiation, such as the scutum (Esd), scuto-scutellar suture (Ses-es), scutellum (Esl), subscutellum (Sbe), and mediotergite (Mdt). These anatomical features of the thorax are fundamental for the classification of fruit fly species. In this study, these specific structures will be described in detail for the two species under investigation, highlighting those that allow for their differentiation.

For this research, the selection of species to be classified was made following the recommendation of the laboratory expert, who considered the relevance of the population dynamics of \textit{Anastrepha fraterculus} and \textit{Ceratitis capitata}. Initially, a larger number of species was proposed for analysis; however, the complexity associated with separating each species into individual vials would have increased the expert's workload. To ensure accuracy in processing and avoid interference with their tasks, the study was limited to these two species.

The selected species are:
\begin{itemize}
	\item \textit{Anastrepha fraterculus}, known as the South American fruit fly.
	\item \textit{Ceratitis capitata}, known as the Mediterranean fruit fly.
\end{itemize}
\input{contenido/mosca_fruta/taxonomia}

%% file: contenido/mosca_fruta/taxonomia.tex

\begin{figure*}[htbp]
    \centering
    \includegraphics[width=0.8\textwidth]{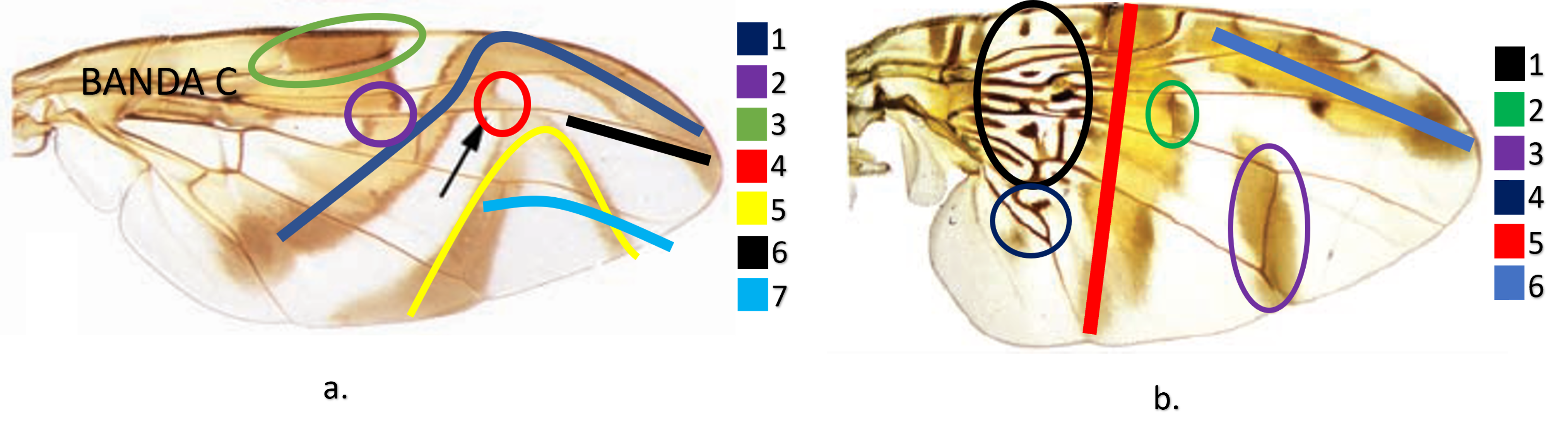}
    \caption{a. Wing of \textit{Anastrepha fraterculus}. b. Wing of \textit{Ceratitis capitata}.}
    \scriptsize \textit{Source:} \cite{especiesImEco}.
    \label{fig:alas}
\end{figure*}

\begin{figure*}[htbp]
    \centering
    \includegraphics[width=0.8\textwidth]{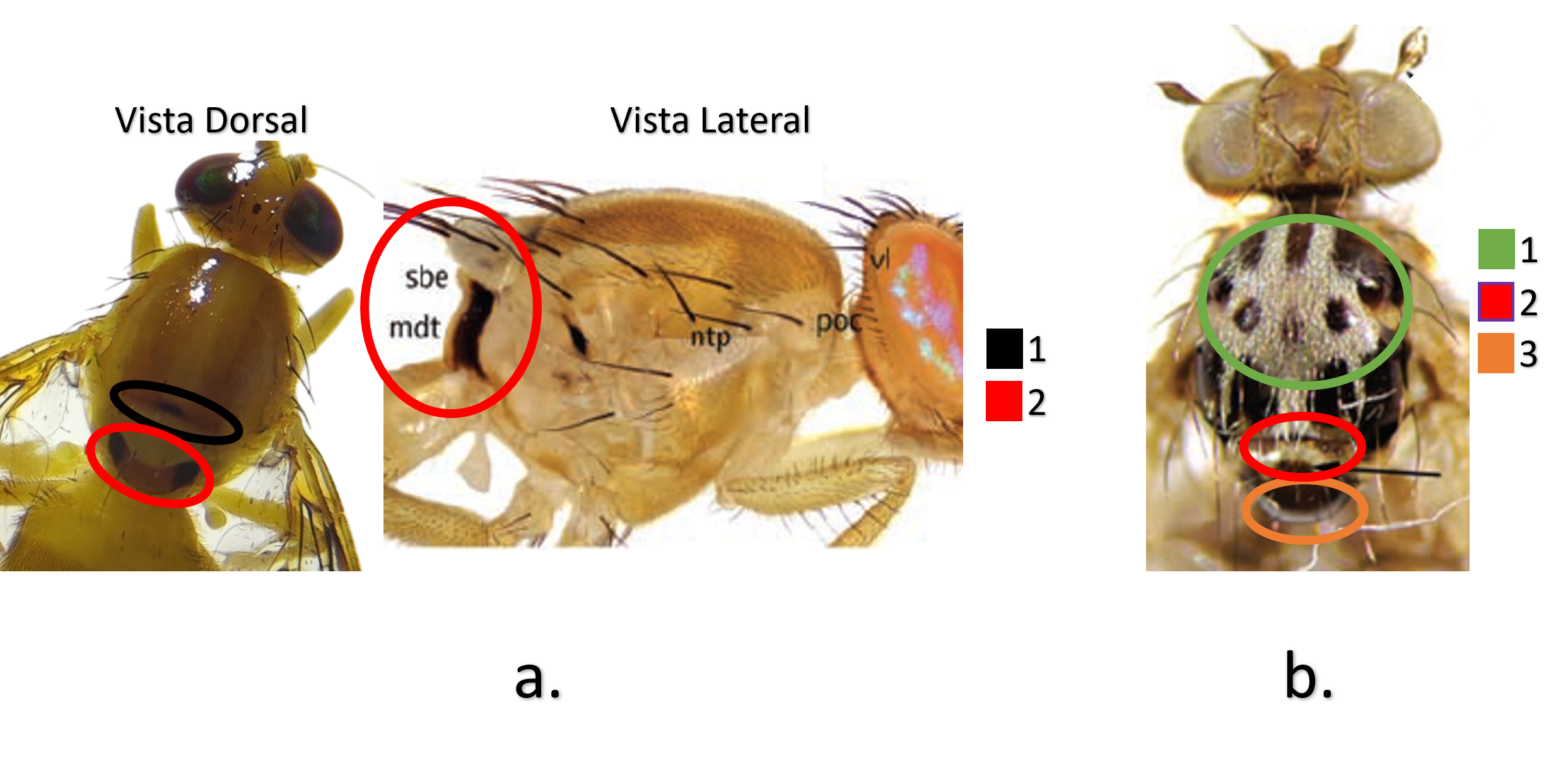}
    \caption{a. Thorax of \textit{Anastrepha fraterculus} b. Thorax of \textit{Ceratitis capitata}.}
    \scriptsize \textit{Source:} \cite{especiesImEco}
    \label{fig:toraxs}
\end{figure*}

Knowing the sections of the wings and the parts of the thorax, we will proceed to identify the morphological characteristics that differentiate the two selected species. These differences can be observed in Figures \ref{fig:alas} and \ref{fig:toraxs}, and are described in Tables \ref{tab:diferenciasalas} and \ref{tab:diferenciastorax}.

\begin{table*}[htbp]
	\centering
	\caption{Differences in wings. \cite{senasaTephritidae, senasaAnastrepha, especiesImEco}}
	\label{tab:diferenciasalas}
	\begin{tabular}{|p{0.4\textwidth}|p{0.05\textwidth}|p{0.4\textwidth}|p{0.05\textwidth}|}
		\hline
		\textbf{\textit{Anastrepha fraterculus}} & \textbf{Figure} & \textbf{\textit{Ceratitis capitata}} & \textbf{Figure} \\
		\hline
		Complete and slightly wide S-band at its apical portion & \ref{fig:alas}.a.1 & Wing with a pattern of yellow stripes & \ref{fig:alas}.b \\
		\hline
		C and S bands fully connected & \ref{fig:alas}.a.2 & Wing with dark spots in cells bc, c, br, bm, and bcu & \ref{fig:alas}.b.1 \\
		\hline
		Hyaline spot at the apex of vein $R_1$ & \ref{fig:alas}.a.3 & r-m vein near the middle of the dm cell within the discal band including the pterostigma & \ref{fig:alas}.b.2 \\
		\hline
		S and V bands generally connected, sometimes with a slight separation & \ref{fig:alas}.a.4 & Costal band pigmenting vein dm-cu does not cross the $r_{4+5}$ cell & \ref{fig:alas}.b.3 \\
		\hline
		Complete V band at its upper portion & \ref{fig:alas}.a.5 & Posteroapical lobe of cell bcu thinner at the base than in the middle & \ref{fig:alas}.b.4 \\
		\hline
		$R_{4+5}$ vein almost straight & \ref{fig:alas}.a.6 & Discal stripe & \ref{fig:alas}.b.5 \\
		\hline
		Moderate apical curvature of vein M & \ref{fig:alas}.a.7 & Costal stripe extended to the apical margin of the wing & \ref{fig:alas}.b.6 \\
		\hline
	\end{tabular}
\end{table*}

\begin{table*}[htbp]
	\centering
	\caption{Differences in the thorax. \cite{senasaTephritidae, senasaAnastrepha, especiesImEco}}
	\label{tab:diferenciastorax}
	\begin{tabular}{|p{0.4\textwidth}|p{0.05\textwidth}|p{0.4\textwidth}|p{0.05\textwidth}|}
		\hline
		\textbf{\textit{Anastrepha fraterculus}} & \textbf{Figure} & \textbf{\textit{Ceratitis capitata}} & \textbf{Figure} \\
		\hline
		Scuto-scutellar suture (Ses-es) mark generally present and extending to the sides & \ref{fig:toraxs}.a.1 & Scutum with irregular yellow spots & \ref{fig:toraxs}.b.1 \\
		\hline
		Scutellum (sbe) with a black spot on each side extending to the mediotergite (mdt) & \ref{fig:toraxs}.a.2 & Yellow mark near the scuto-scutellar suture & \ref{fig:toraxs}.b.2 \\
		\hline
		& & Globose or swollen scutellum with a shiny black color & \ref{fig:toraxs}.b.3 \\
		\hline
	\end{tabular}
\end{table*}

%% file: contenido/materiales_metodos/main_mat_met.tex
\section{Materials and Methods}
\input{contenido/materiales_metodos/adquisicion_imagenes}
\input{contenido/materiales_metodos/preprocesamiento}
\input{contenido/materiales_metodos/transferencia_aprendizaje}
\input{contenido/materiales_metodos/aumento_datos}
\input{contenido/materiales_metodos/grad-cam}

%% file: contenido/materiales_metodos/adquisicion_imagenes.tex
\subsection{Image Acquisition}
Two main categories are considered for the image capture setup: laboratory environments and field environments \cite{18Martineau2017}.

Since SENASA is the competent authority in the identification of fruit fly species, access was requested to one of their facilities. This allowed image capture in a laboratory environment, which follows a standard protocol for capturing, identifying, classifying, and reporting species.

The laboratory protocol is as follows:
\begin{itemize}
    \item Official SENASA McPhail traps (Figure \ref{fig:multilure}) are placed in different districts the province of "La Convención".
        \begin{figure}[htbp]
            \centering
            \includegraphics[width=0.2\textwidth]{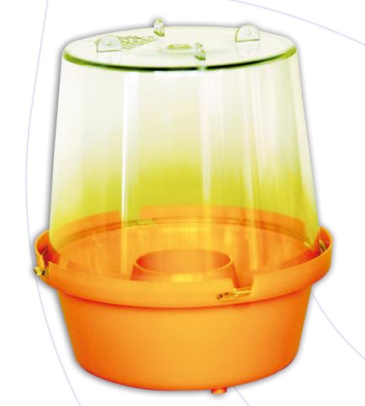}
            \caption{McPhail trap.}
            \scriptsize \textit{Source:} \cite{Safer2022}.
            \label{fig:multilure}
        \end{figure}
    \item The fruit fly species are attracted to the traps using a food attractant made from Torula yeast. The flies that land on this attractant become stuck and trapped.
    \item To collect the flies, the food attractant, along with the trapped flies, is poured into jars labeled according to the district, zone, and specific trap where they were found.
    \item The jars with the samples are handed over to the expert responsible for the identification and classification of the species.
    \item The information about the identified and classified species is recorded in an Official Trap Record (ROT) format, which allows tracking of the traps installed nationwide \cite{6senasa_manual_vigilancia}.
\end{itemize}

During this stage, a dataset of images corresponding to the species \textit{Anastrepha fraterculus} and \textit{Ceratitis capitata} was requested from the entity. However, it was reported that they did not have a dataset with the required characteristics to meet the objectives of this research. Due to this limitation, the creation of a custom dataset was proposed, specifically designed to meet the needs of the study.

In collaboration with the expert responsible for classifying the flies from the traps, the separation of the flies belonging to the species \textit{Anastrepha fraterculus} and \textit{Ceratitis capitata} was coordinated.

Since the essential morphological characteristics for the classification of these species are the wings and thorax, the optimal position to capture the most information from these two structures was identified. After a detailed analysis, it was concluded that the dorsal position provides the maximum visibility of the relevant coloration and sections. However, during the collection process, issues arose with some samples because the flies, when trapped, had their morphological features altered due to the position in which they were caught. For this reason, each sample was carefully handled.

\begin{figure}[htbp]
    \centering
    \includegraphics[width=0.4\textwidth]{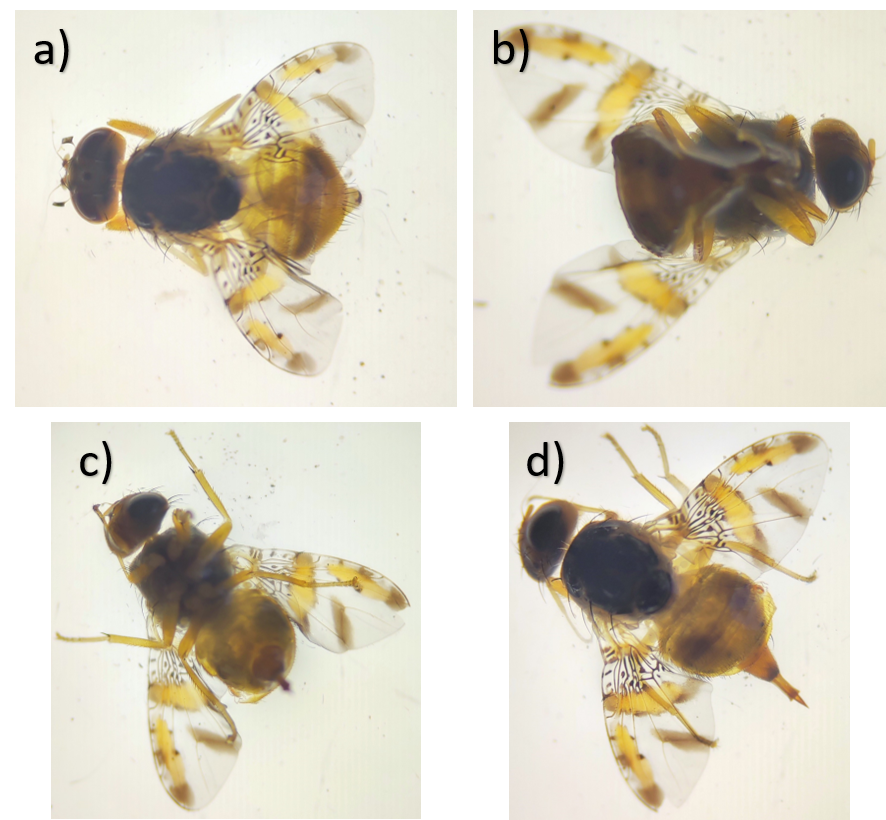}
    \caption{Fly Handling.}
    \scriptsize \textit{Source:} Author's own creation. 
    \label{fig:manipulacion}
\end{figure}

\textbf{Sample Handling.}
To ensure the accuracy of photographic captures of the samples, it is essential to handle each fly in a way that maximizes the visibility of the areas of interest for the correct classification of each species. The following steps were taken to adjust the position of the flies during the handling process:

\begin{enumerate}
	\item First, the fly is observed in the dorsal position (Figure \ref{fig:manipulacion}-a). In this position, the left wing is retracted, preventing full visualization of the area of interest required for classification.
	\item Next, the fly is shown in the ventral position (Figure \ref{fig:manipulacion}-b). In this view, the abdomen is contracted, and the legs are retracted towards the thorax, which hinders proper observation from the dorsal position.
	\item With the fly in the ventral position, the abdomen is manipulated to expand it, and the legs are adjusted to extend from the thorax (Figure \ref{fig:manipulacion}-c). This adjustment allows for better visualization of the relevant structures.
	\item Finally, the fly is returned to the dorsal position (Figure \ref{fig:manipulacion}-d). In this position, the wings are fully extended, unlike what was observed in Figure \ref{fig:manipulacion}-a. Additionally, at the end of the abdomen, the ovipositor is visible, which was not visible in Figure \ref{fig:manipulacion}-a.
\end{enumerate}

\begin{figure}[htbp]
    \centering
    \includegraphics[width=0.3\textwidth]{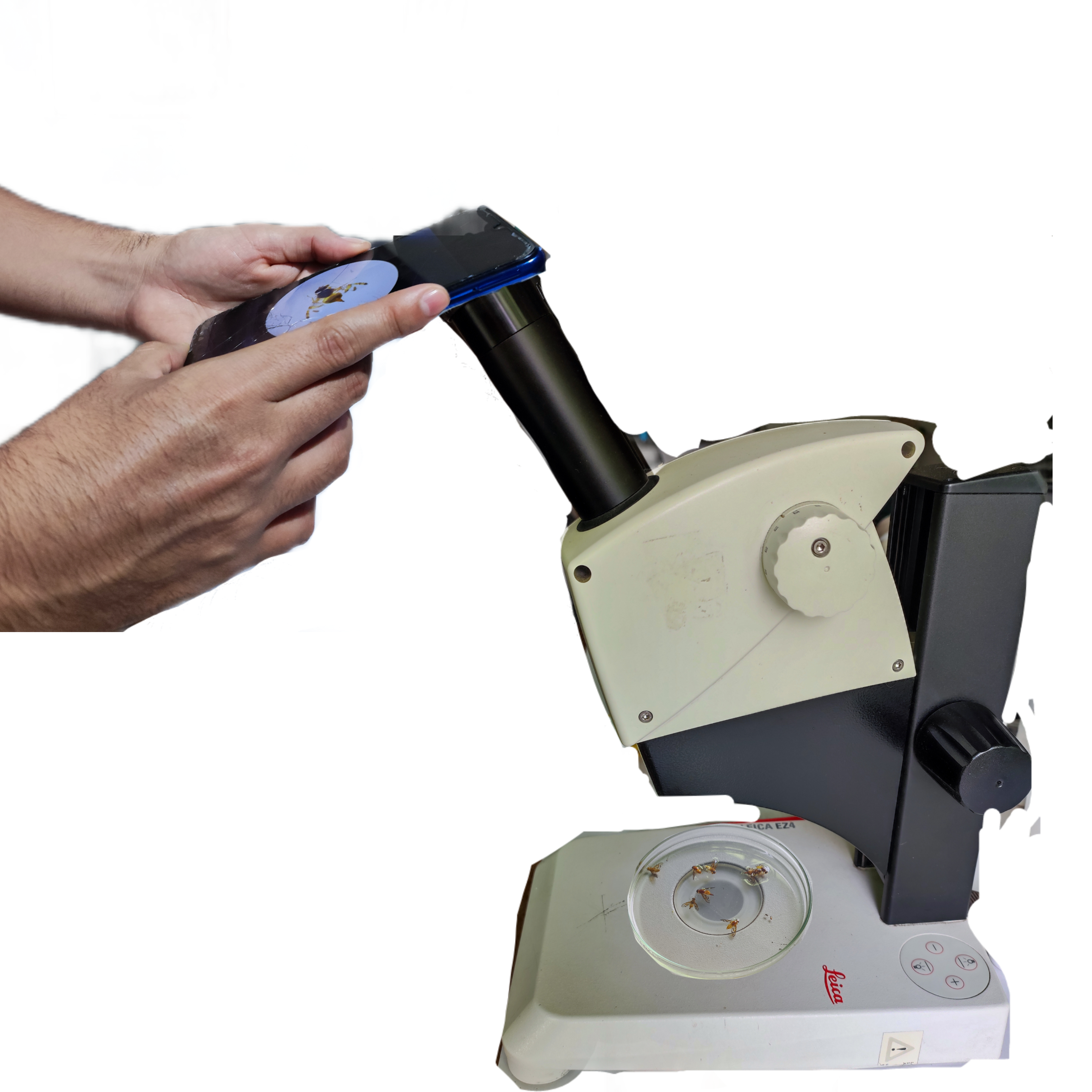}
    \caption{Reference image of the image capture process.}
    \scriptsize \textit{Source:} Author's own creation.
    \label{fig:refcapturaimagen}
\end{figure}

Subsequently, the process of capturing photographs of the species \textit{Anastrepha fraterculus} and \textit{Ceratitis capitata} was carried out. The images were taken in the dorsal position using the camera of a Xiaomi Mi 11 Lite 5G NE smartphone, equipped with autofocus. The captures were made through the ocular lenses of the Leica EZ4 stereomicroscope, as shown in Figure \ref{fig:refcapturaimagen}.

The captured images were stored in JPG format, with resolutions ranging from 2088 x 4640 pixels to 3472 x 4640 pixels. A total of 689 images of \textit{Anastrepha fraterculus} and 286 images of \textit{Ceratitis capitata} were obtained. To balance the number of images per class, 286 images were randomly selected from the \textit{Anastrepha fraterculus} set, thus equalizing the number of images for both species. This resulted in a balanced dataset of 572 images in total.

It is important to emphasize the privacy of this dataset. The obtained images are authored by SENASA and have been personally collected by me for this research. Access to these images is strictly limited to the involved parties, ensuring their controlled and protected use in accordance with the confidentiality and security principles established in the research.
 

%% file: contenido/materiales_metodos/preprocesamiento.tex
\subsection{Preprocessing of the dataset.}
\begin{figure}[htbp]
    \centering
    \includegraphics[width=0.4\textwidth]{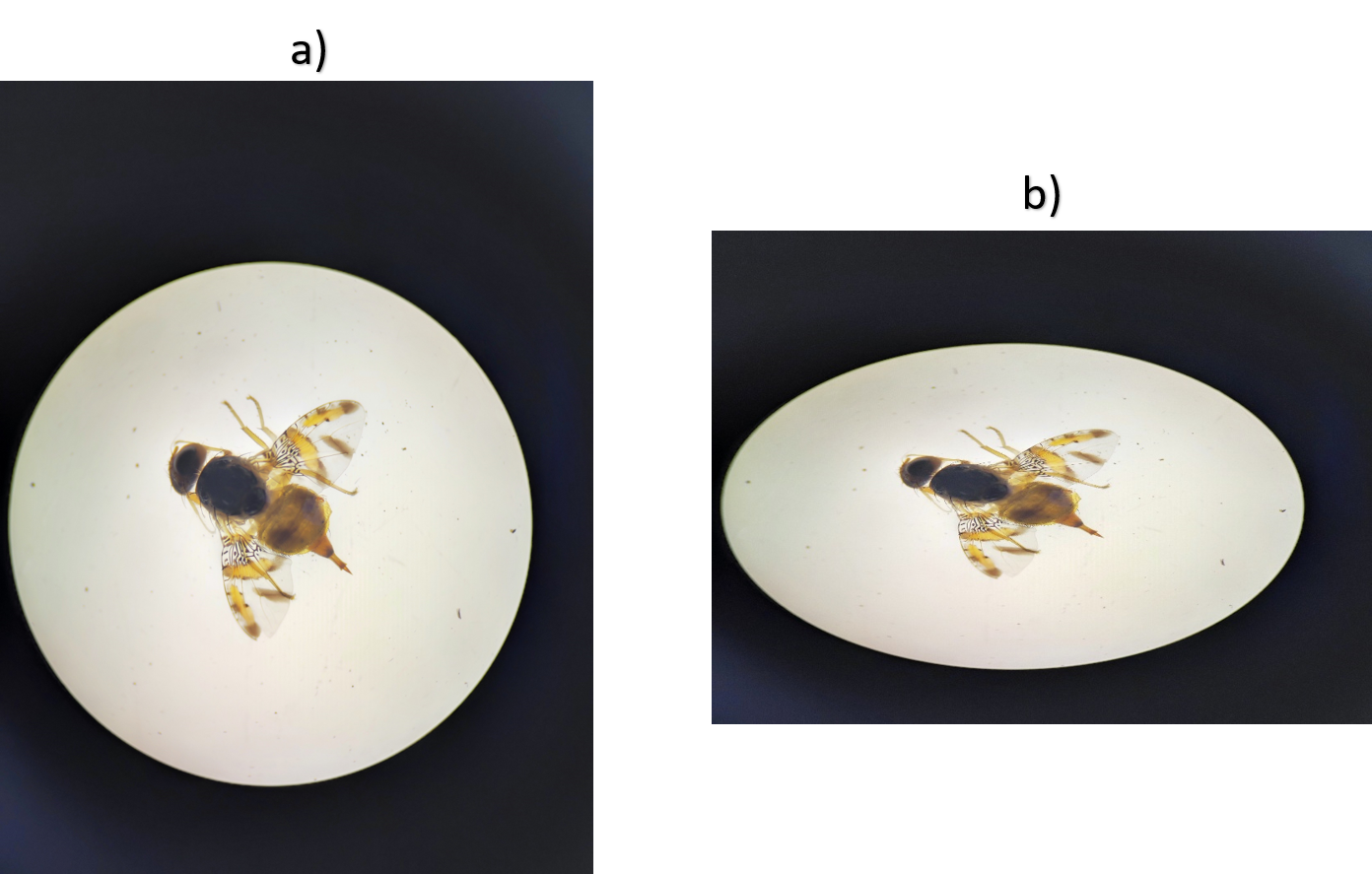}
    \caption{a) Original image. b) 800 x 600 image.}
    \scriptsize \textit{Source:} Author's own creation
    \label{fig:redimension}
\end{figure}

To standardize the image size and facilitate analysis, they were transformed to a resolution of 800 x 600 pixels. This resizing process is crucial to ensure the uniformity of the dataset, and this resizing is shown in Figure \ref{fig:redimension}.

To segment the fly within the image, three preprocessing stages were carried out: segmentation, cropping, and centering. These stages were implemented using the OpenCV library\footnote{OpenCV – 4.9.0 is a library that offers advanced tools for solving computer vision problems through image and video processing.}. Each stage contributes to improving the accuracy and effectiveness of the subsequent analysis by appropriately preparing the images.

\textbf{Fly Segmentation.}
To train our models, we only need the fly area in the images. The following steps were used to segment the flies:
\begin{enumerate}
	\item The image was loaded.
	\item The image was resized to a standard size of 800 x 600 pixels using bilinear interpolation, which ensures good quality in resizing, as illustrated in Figure \ref{fig:redimension}-b.
	\item A median filter was applied to reduce "salt-and-pepper" noise and smooth the image, as shown in Figure \ref{fig:preprocesamiento}-1.
	\item The image was converted to the HSV color space, and a mask was created based on pixel selection within a specific range, as seen in Figures \ref{fig:preprocesamiento}-2 and \ref{fig:preprocesamiento}-3.
	\item The following morphological transformations were applied to the mask to improve segmentation:
	\begin{itemize}
		\item \textbf{Opening.} This operation removed smaller areas that did not contain relevant information, as shown in Figure \ref{fig:preprocesamiento}-4.
		\item \textbf{Dilation.} The selected area was expanded to include nearby relevant information, as seen in Figure \ref{fig:preprocesamiento}-5.
		\item \textbf{Closing.} This operation filled small areas within the regions identified by opening and dilation that were not previously selected, as observed in Figure \ref{fig:preprocesamiento}-6.
	\end{itemize}
	\begin{figure}[htbp]
		\centering
		\includegraphics[width=0.4\textwidth]{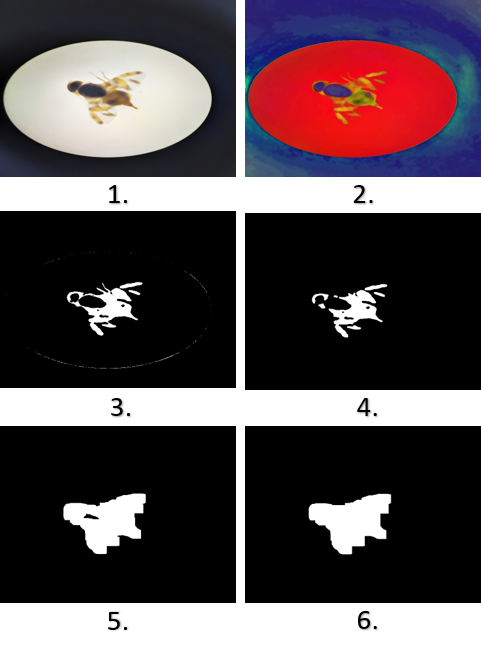}
		\caption{Steps of image preprocessing.}
		\scriptsize \textit{Source:} Author's own creation
		\label{fig:preprocesamiento}
	\end{figure}
	\item The mask was combined with the original image to visualize the processed area, and the pixels with a value of 0 were changed to 255 to set the background in white, as shown in Figure \ref{fig:imagenfinal}.
	\begin{figure}[htbp]
		\centering
		\includegraphics[width=0.2\textwidth]{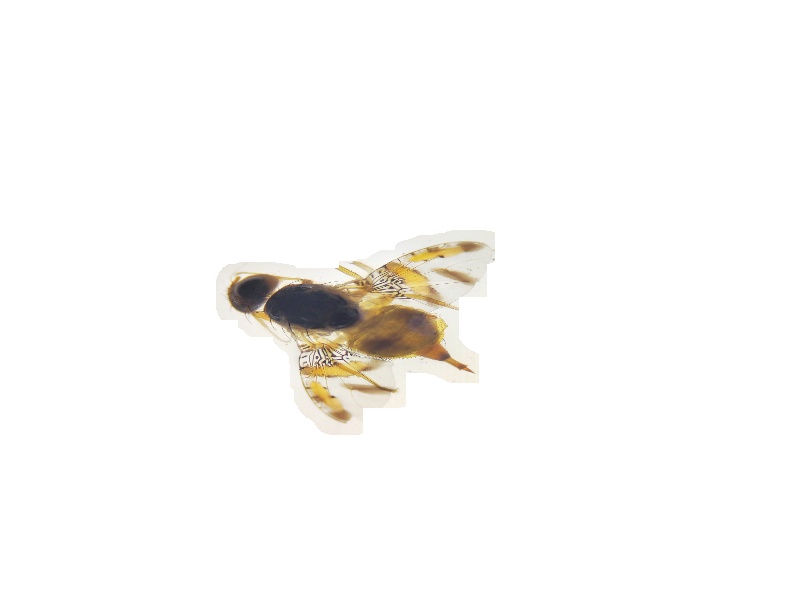}
		\caption{Final preprocessed image.}
		\scriptsize \textit{Source:} Author's own creation
		\label{fig:imagenfinal}
	\end{figure}
\end{enumerate}

\begin{figure}[htbp]
    \centering
    \includegraphics[width=0.3\textwidth]{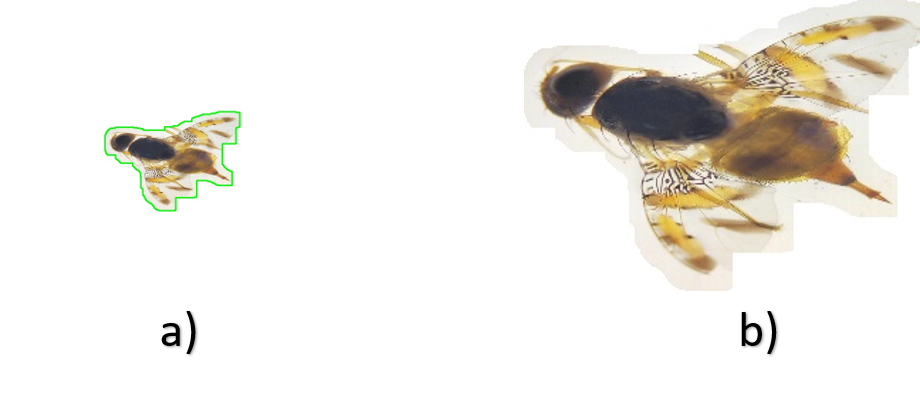}
    \caption{a) Largest contour found. b) Cropped image with coordinates of the largest contour.}
    \label{fig:recorte}
    \scriptsize \textit{Source:} Author's own creation
\end{figure}

\textbf{Cropping.}
To focus the image and eliminate the irrelevant background, the cropped image was processed. The cropping process was carried out as follows:
\begin{enumerate}
	\item Using the mask obtained in the previous step, the largest contour in the segmented image was detected. The coordinates of this contour were identified and used to define the region to crop, as shown in Figure \ref{fig:recorte}-a.
	\item With the coordinates of the largest contour, the image was cropped, removing the background and retaining only the region of interest, as shown in Figure \ref{fig:recorte}-b.
\end{enumerate}

\textbf{Centering.}
Once the image was segmented and cropped, it was resized to a uniform format. A size of 400 x 400 pixels was chosen, and the missing pixels of each image were filled to reach the desired dimensions, ensuring the image was centered within that space.
\begin{enumerate}
	\item A target size of 400 x 400 pixels was set for each image to avoid excessive reduction of images. 
	\item A scaling factor was determined as the quotient between the desired size and the largest value between the height and width of the original image. This ensures that the image is resized proportionally without distorting its dimensions.
	\item The image was resized using the calculated scaling factor, thus maintaining the original proportions. The resizing was performed using the cv2.resize function from OpenCV.
	\item The resized image was placed in a new 400 x 400 pixel image. The remaining area was filled with a white background to ensure the image maintained the desired uniform size. This was done using the cv2.copyMakeBorder function from OpenCV, which adds a white border around the resized image.
	\item The centered image was saved to a file for later use.
\end{enumerate}

\begin{figure}[htbp]
    \centering
    \includegraphics[width=0.2\textwidth]{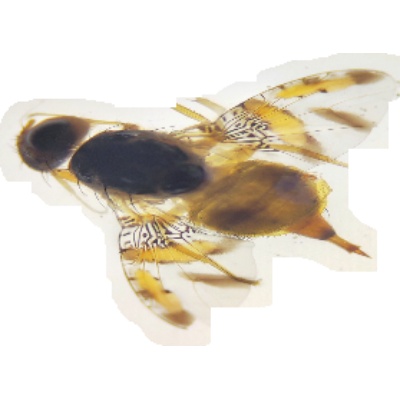}
    \caption{Segmented and centered image.}
    \scriptsize \textit{Source:} Author's own creation
    \label{fig:centrada}
\end{figure}

As a result, an image was obtained that, after segmentation, cropping, and centering, has a uniform size of 400 x 400 pixels, as shown in Figure \ref{fig:centrada}.

%% file: contenido/materiales_metodos/transferencia_aprendizaje.tex
\subsection{Transfer Learning.}
\label{subsec:Transfer}

In the context of this research, the pre-trained models VGG16 and VGG19 will be used, as they have demonstrated to be the best feature extractors in the study by \cite{12Leonardo2018}. Furthermore, the Inception-V1 model showed promising results in \cite{15Gonzalez-Lopez2022}, where the author recommends using more advanced versions of this model, as they may offer further improvements. Based on this recommendation, Inception-V3 will be employed. This model represents an enhanced iteration of Inception-V1, with improvements in accuracy and efficiency.

%% file: contenido/materiales_metodos/aumento_datos.tex
\subsection{Data Augmentation.}

Given the need for large amounts of training data, the collection and labeling process being slow, complex, and costly \cite{22Toke2020}, it is imperative to use techniques that allow for model training with smaller datasets. In this context, data augmentation presents itself as an effective solution in the field of deep learning.

\begin{figure}[htbp]
    \centering
    \includegraphics[width=0.4\textwidth]{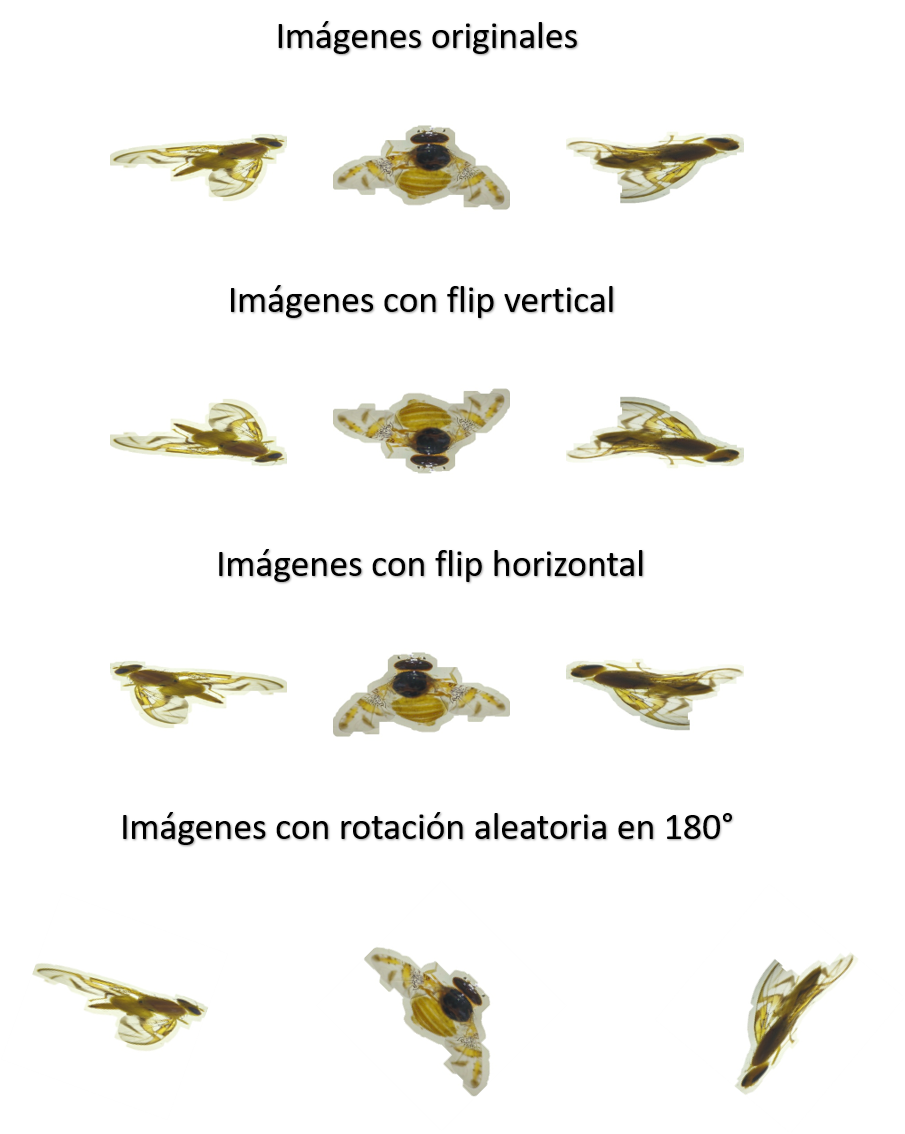}
    \caption{Example of data augmentation.}
    \scriptsize \textit{Source:} Author's own creation
    \label{fig:aumento_datos}
\end{figure}

For our training dataset, we will apply basic image manipulation methods such as random rotation, horizontal and vertical flipping, as shown in Figure \ref{fig:aumento_datos}. These techniques have proven to be effective in previous studies \cite{9Martins2019, 14Molina-Rotger2023, 15Gonzalez-Lopez2022, 16Mamdouh2021}, providing significant improvements in training quality and model performance.

%% file: contenido/materiales_metodos/grad-cam.tex
\subsection{Model Explainability.}

In Section \ref{sec:moscafruta}, we explain the morphological features that experts identify when classifying the species \textit{Anastrepha fraterculus} and \textit{Ceratitis capitata}.

We need to be able to explain the predictions made by our models, which is why we will use the Grad-Cam technique. This method allows us to represent what our model is "looking at" through the last convolutional layer.

This will allow us to verify that our models are correctly identifying the morphological features of each species \cite{24Selvaraju2017}.

%% file: contenido/experimentos_resultados/main_exp_res.tex
\section{Experiments and Results}

This section details the experiments conducted and the results obtained in our work. For this, three selected models were employed: VGG16, VGG19, and Inception-V3. The specific results of each model are presented below, along with a comparative analysis of their performance, evaluated in terms of accuracy, recall, F1-score, and classification time.

\input{contenido/experimentos_resultados/segmentacion}
\input{contenido/experimentos_resultados/entrenamiento}

\input{contenido/experimentos_resultados/resultados}

%% file: contenido/experimentos_resultados/segmentacion.tex
\subsection{Segmentation.}

\begin{figure}[htbp]
    \centering
    \includegraphics[width=0.4\textwidth]{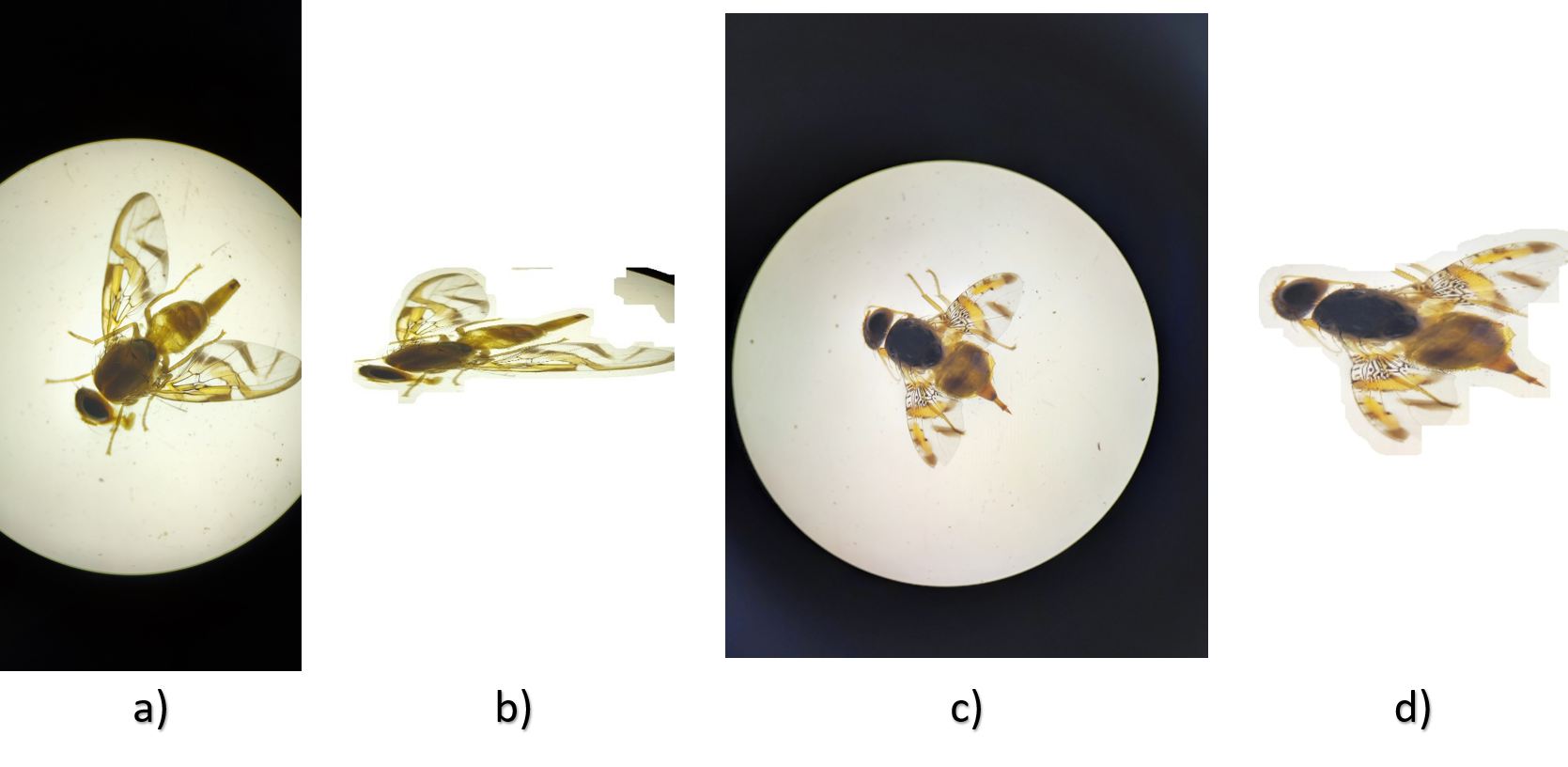}
    \caption{a) Original image. b) Incorrectly segmented image. c) Original image. d) Correctly segmented image.}
    \scriptsize \textit{Source:} Author's own creation
    \label{fig:segmentacioncorrecta}
\end{figure}

The dataset was generated from the manual capture of images using the eyepieces of a stereomicroscope, which resulted in variations in contrast, lighting, and brightness of the images. During the segmentation process, it was observed that some of these images were not correctly segmented. In Figure \ref{fig:segmentacioncorrecta}.b, which shows the segmentation of Figure \ref{fig:segmentacioncorrecta}.a, it can be seen that the image retains parts of the background, which are considered noise. This noise could affect the predictions of the models, preventing them from being properly trained. On the other hand, Figure \ref{fig:segmentacioncorrecta}.d shows correct segmentation, without background remnants or noise, ensuring higher prediction accuracy.

To ensure proper training of the models, images that were not correctly segmented were manually separated. These included 50 discarded images per class, which reduced the dataset size to 236 images per class.

%% file: contenido/experimentos_resultados/entrenamiento.tex
\subsection{Training.}

As described in \cite{20Simonyan2015}, the VGG16 and VGG19 models were originally trained with images of 224 x 224 pixels. On the other hand, \cite{21Szegedy2016} mentions that the Inception-v3 model was trained with images of 299 x 299 pixels, although it can also process images of lower resolution. While training with lower-resolution images may increase the required time, the quality of the final result is quite close to that obtained with the original resolution.

Based on the information provided by \cite{20Simonyan2015, 21Szegedy2016}, we decided to resize our dataset to images of 224 x 224 pixels and three color channels (224 x 224 x 3), maintaining input data homogeneity for the three models. This way, we maintain the proper resolution for the VGG16 and VGG19 models and ensure acceptable performance for the Inception-v3 model. Additionally, we used the specific preprocessing functions for each model to standardize the inputs.

The \textit{preprocess\_input()} function of VGG16 \cite{25tensorflowvgg16} and VGG19 \cite{26tensorflowvgg17} converts the image array to the \textit{float32} format, changes the color space from RGB to BGR, and performs zero-centering, i.e., subtracts the mean of each color channel from the corresponding pixel values.

In the case of Inception-v3, the \textit{preprocess\_input()} function \cite{27tensorflowinceptionv3} also converts the image array to \textit{float32}, but additionally scales the pixel values in a range from -1 to 1.

\textbf{Dataset Split.} The dataset used in this study consists of a total of 472 images. It was divided into three subsets: 70\% for training, 20\% for validation, and 10\% for testing. The images that had incorrect segmentations were included in the test set in order to evaluate the model's ability to generalize and correctly classify both well-segmented images and those with segmentation errors. This detailed split is presented in Table \ref{tab:datosentrenamiento}.

\begin{table}[htbp]
	\centering
	\caption{Dataset Partitioning.}
	\label{tab:datosentrenamiento}
	\begin{tabular}{|l|c|c|c|}
		\hline
		& \textbf{Training} & \textbf{Validation} & \textbf{Test} \\
		\hline
		\textbf{\textit{A. fraterculus}} & 165 & 47 & 24 + 50 \\
		\hline
		\textbf{\textit{C. capitata}} & 165 & 47 & 24 + 50 \\
		\hline
		\textbf{Total} & 330 & 94 & 148 \\
		\hline
	\end{tabular}
\end{table}

\textbf{Hyperparameter Definition.} For each model, we used the following hyperparameters.

\begin{itemize}
	\item Learning rate: 0.0001
	\item Number of epochs: 100
	\item Loss function: categorical\_crossentropy
	\item Optimizer: SGD with momentum of 0.8
	\item Batch size: 16
\end{itemize}

\textbf{Callback Definitions.} In our training approach, we used two callbacks: EarlyStopping and ModelCheckpoint.

\begin{itemize}
	\item   EarlyStopping. 
	This callback stops the training of a model when the selected metric stops improving for a specified number of epochs. This strategy is essential to prevent overfitting the model.
	\item   ModelCheckpoint. 
	This callback saves the model at periodic checkpoints, either after each epoch or when a new best performance on a specific metric is reached. This allows us to preserve the best model obtained during training.
\end{itemize}

By combining these two callbacks in our training strategy, we achieved greater robustness and efficiency in the training process of our models. We prevent overfitting by stopping training when no improvement in a specific metric is observed over a period, while simultaneously allowing us to preserve the best model obtained so far.

For our research, we set the number of epochs for EarlyStopping to 25 and "val\_accuracy" as the monitored metric for both callbacks.

%% file: contenido/experimentos_resultados/resultados.tex
\subsection{Results.}
After running experiments with each selected model, the following results were obtained:

\textbf{VGG16 Model.} The best performance for this model was achieved in epoch 91, with a total training time of 4750.1 seconds. For the test set, consisting of 148 images, the classification time was \textbf{28.88} seconds for \textit{Anastrepha fraterculus} samples and \textbf{16.25} seconds for \textit{Ceratitis capitata} samples, with a total classification time of \textbf{45.13} seconds.

\textbf{VGG19 Model.} The VGG19 model achieved its best performance in epoch 79, with a total training time of 5831.29 seconds. In the test set of 148 images, the classification time was \textbf{16.81} seconds for \textit{Anastrepha fraterculus} samples and \textbf{24.02} seconds for \textit{Ceratitis capitata} samples, with a total classification time of \textbf{40.83} seconds.

\textbf{Inception-v3 Model.} The best performance for the Inception-v3 model was achieved in epoch 41, with a total training time of 806.56 seconds. In the test set of 148 images, the classification time was \textbf{7.68} seconds for \textit{Anastrepha fraterculus} samples and \textbf{7.36} seconds for \textit{Ceratitis capitata} samples, with a total classification time of \textbf{15.04} seconds.

Table \ref{tab:resultados_experimentos} presents the confusion matrices corresponding to the classification performance of VGG16, VGG19, and Inception-V3. Each matrix is structured to display the classification results for both species.

\begin{table}[htbp]
	\centering
	\caption{Confusion matrix result. TP = True Positives FP = False Positives FN = False Negatives TN = True Negatives}
	\label{tab:resultados_experimentos}
	\begin{tabular}{|l|c|c|c|c|}
		\hline
		\textbf{MODELS} & \textbf{TP} & \textbf{FN} & \textbf{FP} & \textbf{TN}  \\
		   \hline
		\textbf{Vgg16 \textit{A. fraterculus}} & 50 & 24 & 2 & 72 \\
		\hline
		\textbf{Vgg16 \textit{C. capitata}} & 72 & 2 & 24 & 50 \\
		\hline
		\textbf{Vgg19 \textit{A. fraterculus}} & 53 & 21 & 6 & 68 \\
		\hline
		\textbf{Vgg19 \textit{C. capitata}} & 68 & 6 & 21 & 53 \\
		\hline
		\textbf{Inception-v3 \textit{A. fraterculus}} & 66 & 8 & 3 & 71 \\
		\hline
		\textbf{Inception-v3 \textit{C. capitata}} & 71 & 3 & 8 & 66 \\
		\hline
	\end{tabular}
\end{table}

\begin{figure*}[htb]
	\centering
	\includegraphics[width=1\textwidth]{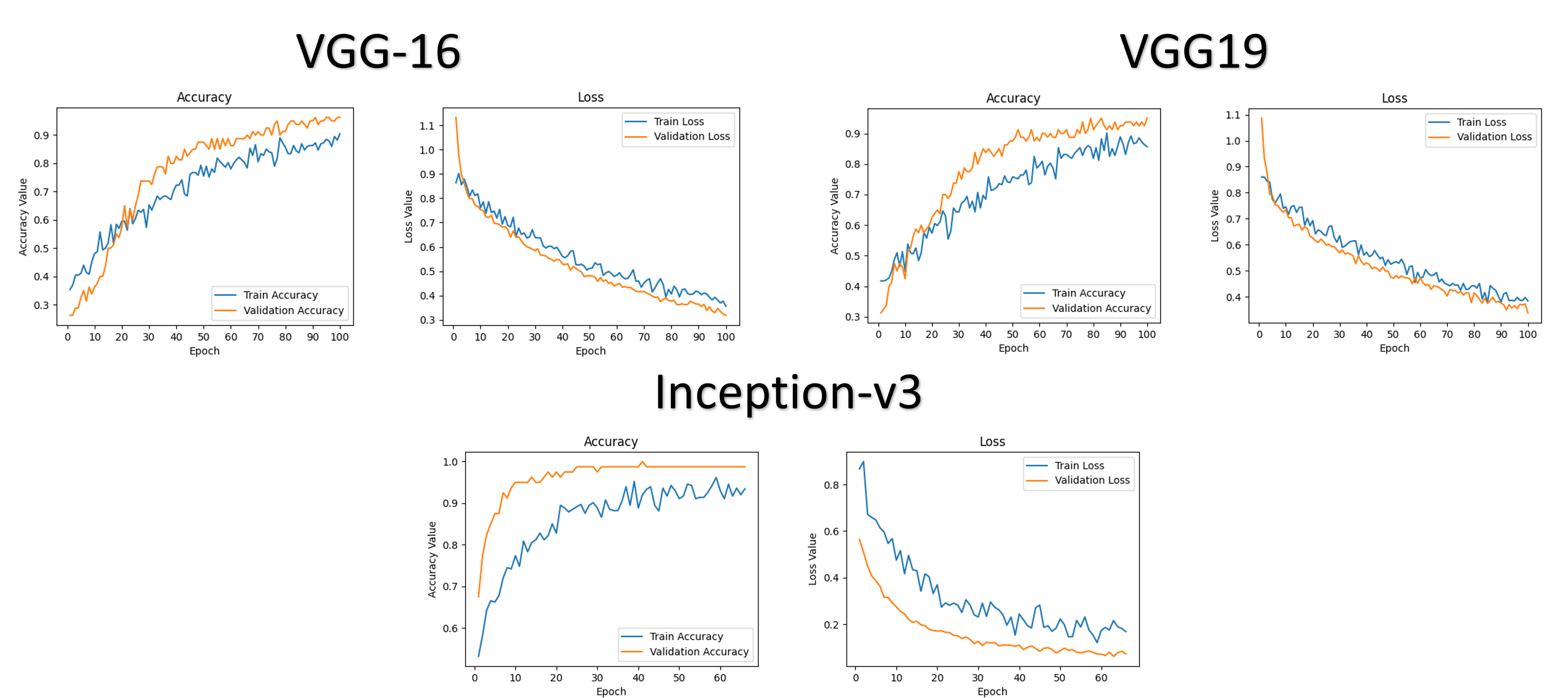}
	\caption{Performance curves of model training.}
	\scriptsize \textit{Source:} Author's own creation
	\label{fig:performance}
\end{figure*}

In Figure \ref{fig:performance}, the performance curves of each model are shown, evaluated through accuracy and loss metrics:

\begin{itemize}
	\item \textbf{Train Accuracy:} Indicates how effectively the models have learned the patterns present in the training dataset.
	\item \textbf{Validation Accuracy:} Reflects how the model's accuracy improves when classifying new, unseen images, evaluating its generalization ability.
	\item \textbf{Train Loss:} Shows how the model adjusts its parameters to reduce errors on the training dataset.
	\item \textbf{Validation Loss:} Measures how the model minimizes errors on the validation data, essential for assessing its performance on unseen data.
\end{itemize}

Analyzing these curves, a similar behavior is observed across all three models. During the initial epochs, the accuracy curves start at low levels. It is notable that the Validation Accuracy curve quickly surpasses the Train Accuracy curve, indicating that the models manage to generalize well from early stages. Subsequently, the Validation Accuracy stabilizes and consistently stays above the Train Accuracy, suggesting that the models did not suffer from overfitting and were effectively trained.

Complementing this analysis, the loss curves also provide relevant insights. Initially, both Train Loss and Validation Loss curves present high values, indicating a significant amount of initial errors. However, as the epochs progress, these curves gradually decrease, showing that the models effectively adjusted their parameters, reducing errors in both the training and validation data. This behavior reinforces the idea that the models did not memorize the training data, avoiding overfitting and correctly generalizing to unseen images.

%% file: contenido/discusion/main_discus.tex
\section{Discusión}

\input{contenido/discusion/segmentacion}
\input{contenido/discusion/resultados}

%% file: contenido/discusion/segmentacion.tex
\subsection{Segmentation}

During the segmentation process, it was identified that some images were affected by various factors associated with the capture process, resulting in a reduction of the original dataset by approximately 17.5\%.

Factors such as brightness, contrast, and lighting conditions were found to have a significant impact on the ability of a segmentation algorithm to generalize correctly. Although manual capture with mobile devices facilitates quick data collection and instant storage, these uncontrolled conditions introduce additional challenges in later stages, such as preprocessing, thereby affecting the quality and consistency of the segmentation.

However, the images discarded due to poor segmentation were not completely removed. They were included in the test set, playing an important role in evaluating the models' ability to generalize and classify images with varied characteristics, including those with imperfect segmentations.

%% file: contenido/discusion/resultados.tex
\subsection{Results}
The classification task in this research does not align with a typical binary classification scenario, where classes are usually defined as "normal" and "abnormal." Instead, we work with two specific classes: \textit{Anastrepha fraterculus} and \textit{Ceratitis capitata}, whose correct classification is crucial in the context of our study.

Since we need to evaluate the classifier's performance for both classes accurately, it is essential to employ metrics that provide detailed information about the model's behavior for each of them. While accuracy and recall are useful, they are not sufficient on their own to correctly assess the model's performance, as we need to consider reducing both false positive and false negative rates. This is especially important due to the critical nature of correctly classifying both species and the meticulous work carried out by SENASA experts to identify the morphological characteristics of each species.

Minimizing false positives and false negatives is essential to ensure accuracy in the reports on the population of each species in a given area. This guarantees the integrity of the results and prevents misinterpretation of the data, which could impact the population analysis of the studied species.

A thorough analysis of our results must be conducted using the F1-score metric, which provides a balance between precision and recall. However, since we need an overall result that reflects the performance across both classes, we use the "macro average," as recommended by the documentation in \cite{scikitlearn}. This approach calculates the average of the metrics for each class, giving equal weight to both, which allows us to obtain a balanced view of the classifier's performance. The metrics calculated with the results from the classification of the test data for the VGG16 model are presented in Table \ref{tab:resultadosvgg16}, for the VGG19 model in Table \ref{tab:resultadosvgg19}, and for the Inception-V3 model in Table \ref{tab:resultadosinception}.
\begin{table}[htbp]
  \centering
  \caption{Metrics of the VGG16 Model}
  \label{tab:resultadosvgg16}
  \begin{tabular}{|l|c|c|c|}
  \hline
     & \textbf{Precision} & \textbf{Recall} & \textbf{F1-score} \\
    \hline
    \textbf{\textit{A. fraterculus}} & 0.96 & 0.68 & 0.79 \\
    \hline
    \textbf{\textit{C. capitata}} & 0.75 & 0.97 & 0.85 \\
    \hline
    \textbf{Macro average} & 0.86 & 0.82 & \textbf{0.82} \\
    \hline
  \end{tabular}
\end{table}

\begin{table}[htbp]
  \centering
  \caption{Metrics of the VGG19 Model}
  \label{tab:resultadosvgg19}
  \begin{tabular}{|l|c|c|c|}
  \hline
     & \textbf{Precision} & \textbf{Recall} & \textbf{F1-score} \\
    \hline
    \textbf{\textit{A. fraterculus}} & 0.90 & 0.72 & 0.80 \\
    \hline
    \textbf{\textit{C. capitata}} & 0.76 & 0.92 & 0.83 \\
    \hline
    \textbf{Macro average} & 0.83 & 0.82 & \textbf{0.82} \\
    \hline
  \end{tabular}
\end{table}

\begin{table}[htbp]
	\centering
	\caption{Metrics of the Inception-v3 Model}
	\label{tab:resultadosinception}
	\begin{tabular}{|l|c|c|c|}
		\hline
		& \textbf{Precision} & \textbf{Recall} & \textbf{F1-score} \\
		\hline
		\textbf{\textit{A. fraterculus}} & 0.96 & 0.89 & 0.92 \\
		\hline
		\textbf{\textit{C. capitata}} & 0.90 & 0.96 & 0.93 \\
		\hline
		\textbf{Macro average} & 0.93 & 0.93 & \textbf{0.93} \\
		\hline
	\end{tabular}
\end{table}

When analyzing the results obtained by our three models, we observed that we achieved good performance in classifying the two species of fruit flies. However, as mentioned in Section \ref{sec:moscafruta}, there are specific morphological characteristics for each species that the models must be able to correctly identify.

\subsection{Analysis of Model Explainability.}
\label{sec:cam}
\begin{figure}[htbp]
    \centering
    \includegraphics[width=0.3\textwidth]{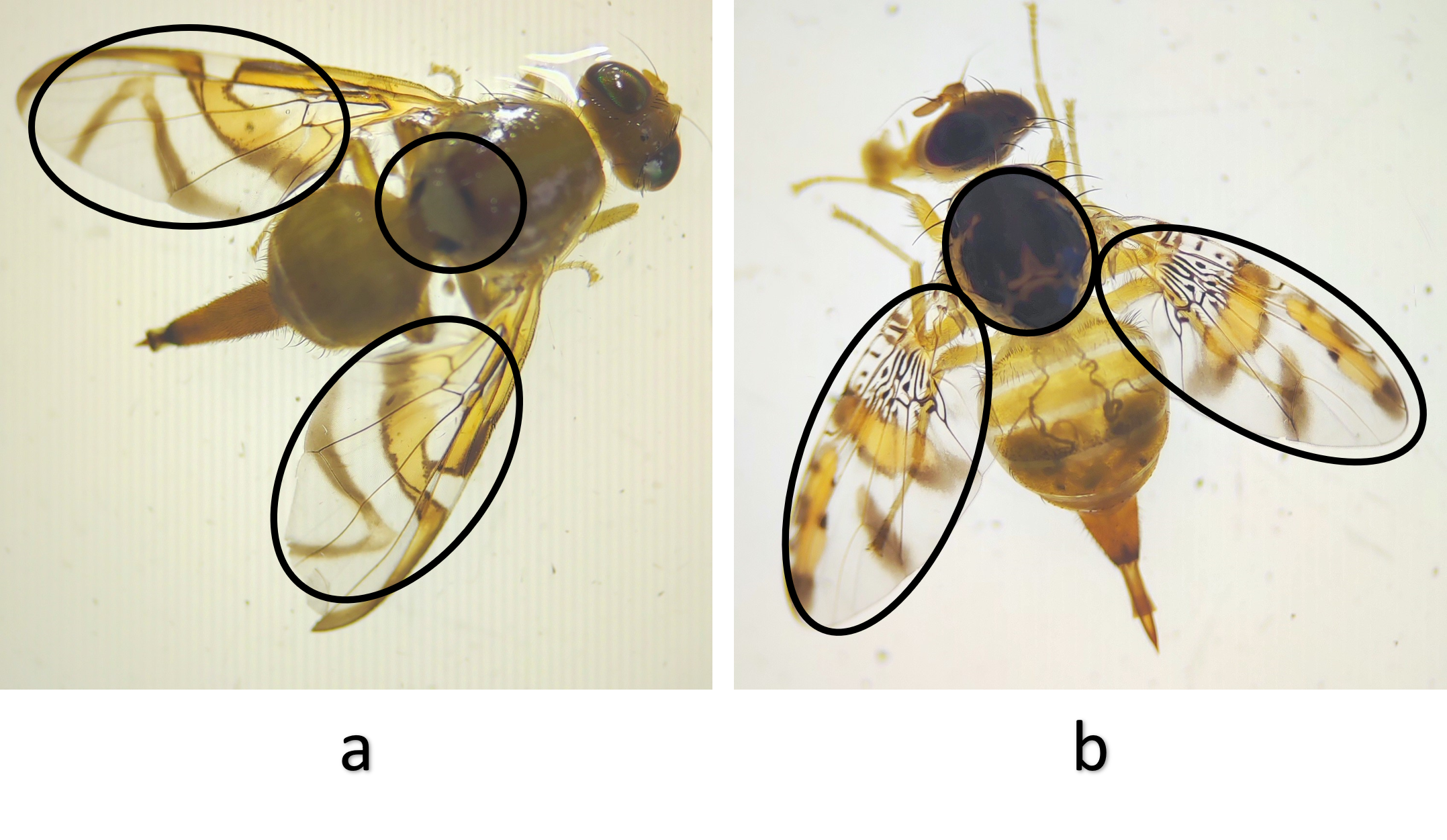}
    \caption{Test images with areas that our models should focus on: a. \textit{Anastrepha fraterculus}. b. \textit{Ceratitis capitata}}
    \scriptsize \textit{Source:} Author's own creation
    \label{fig:moscaprueba}
\end{figure}

To better understand which visual aspects the models are considering when making classifications, we employed the Grad-CAM technique. This technique allows us to generate a visual representation of the areas of the images that the models consider most relevant when assigning them to a specific class. For this, we used two images from our test set, as shown in (Figure \ref{fig:moscaprueba}).

In Section \ref{sec:moscafruta}, we describe the key anatomical sections of fruit flies, such as the thorax, wings, and ovipositor. The ovipositor, which is a crucial feature for classifying these species, is excluded as a morphological feature for classification because it is not possible to accurately differentiate this structure in our images, and it will not be considered an area of interest for our models.

In the images shown in Figure \ref{fig:moscaprueba}.a and \ref{fig:moscaprueba}.b, the areas that should be considered of interest by our models are highlighted with black circles.

\begin{figure}[htbp]
    \centering
    \includegraphics[width=0.4\textwidth]{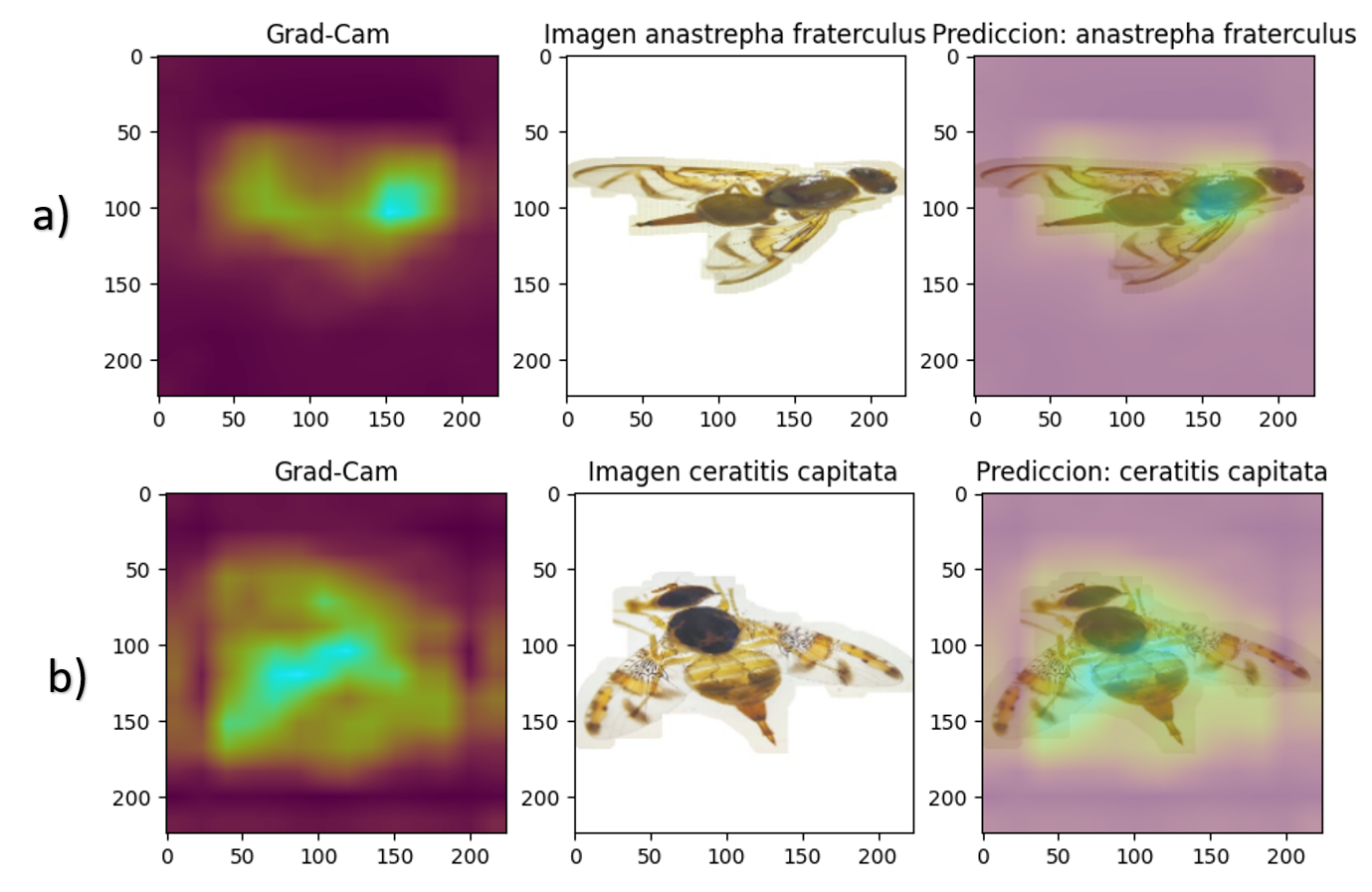}
    \caption{Grad-Cam of the VGG16 model. a) \textit{Anastrepha fraterculus}. b) \textit{Ceratitis capitata}.}
    \scriptsize \textit{Source:} Author's own creation
    \label{fig:camvgg16}
\end{figure}

\begin{figure}[htbp]
    \centering
    \includegraphics[width=0.4\textwidth]{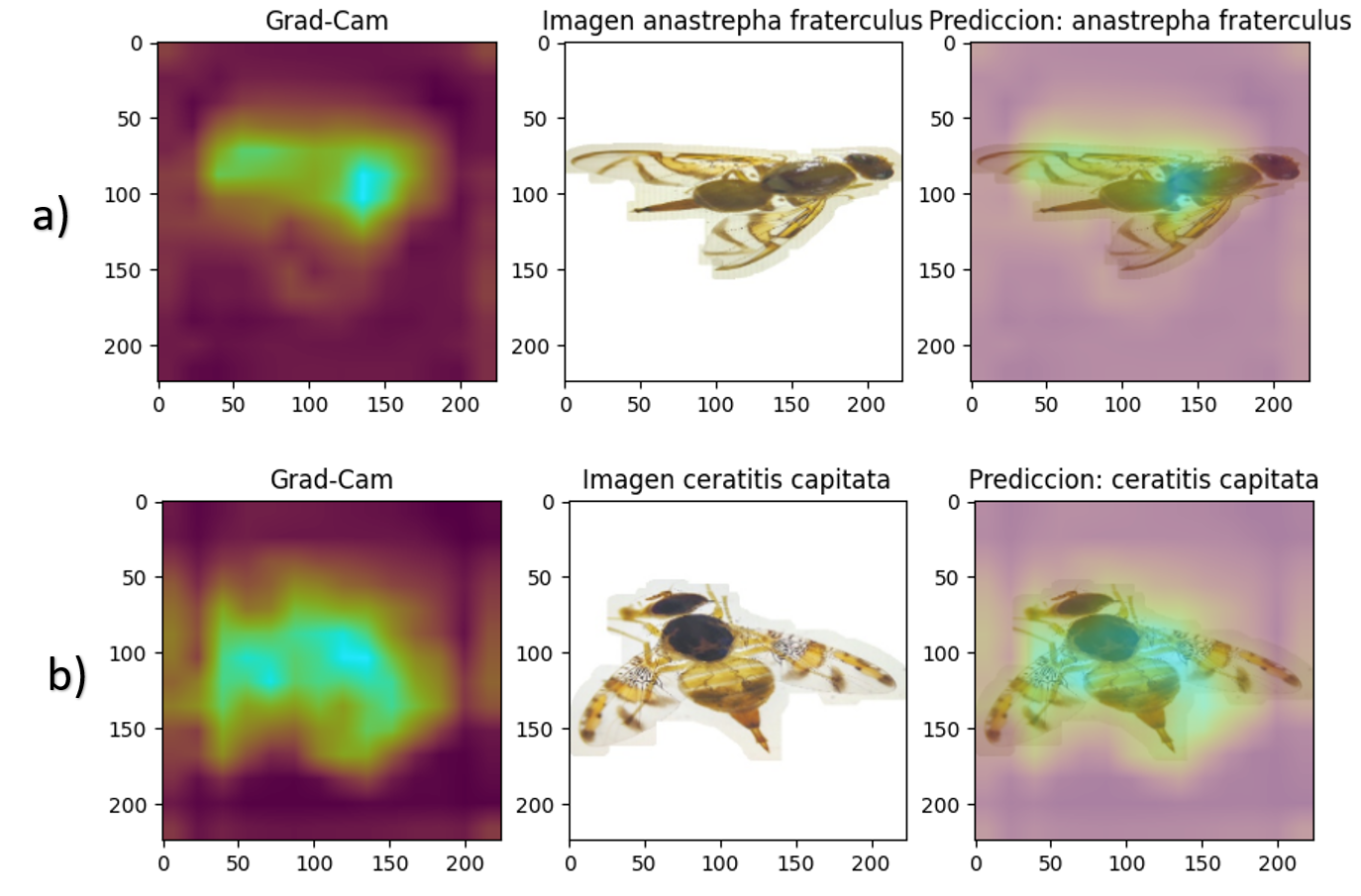}
    \caption{Grad-Cam of the VGG19 model. a) \textit{Anastrepha fraterculus}. b) \textit{Ceratitis capitata}.}
    \scriptsize \textit{Source:} Author's own creation
    \label{fig:camvgg19}
\end{figure}

\begin{figure}[htbp]
    \centering
    \includegraphics[width=0.4\textwidth]{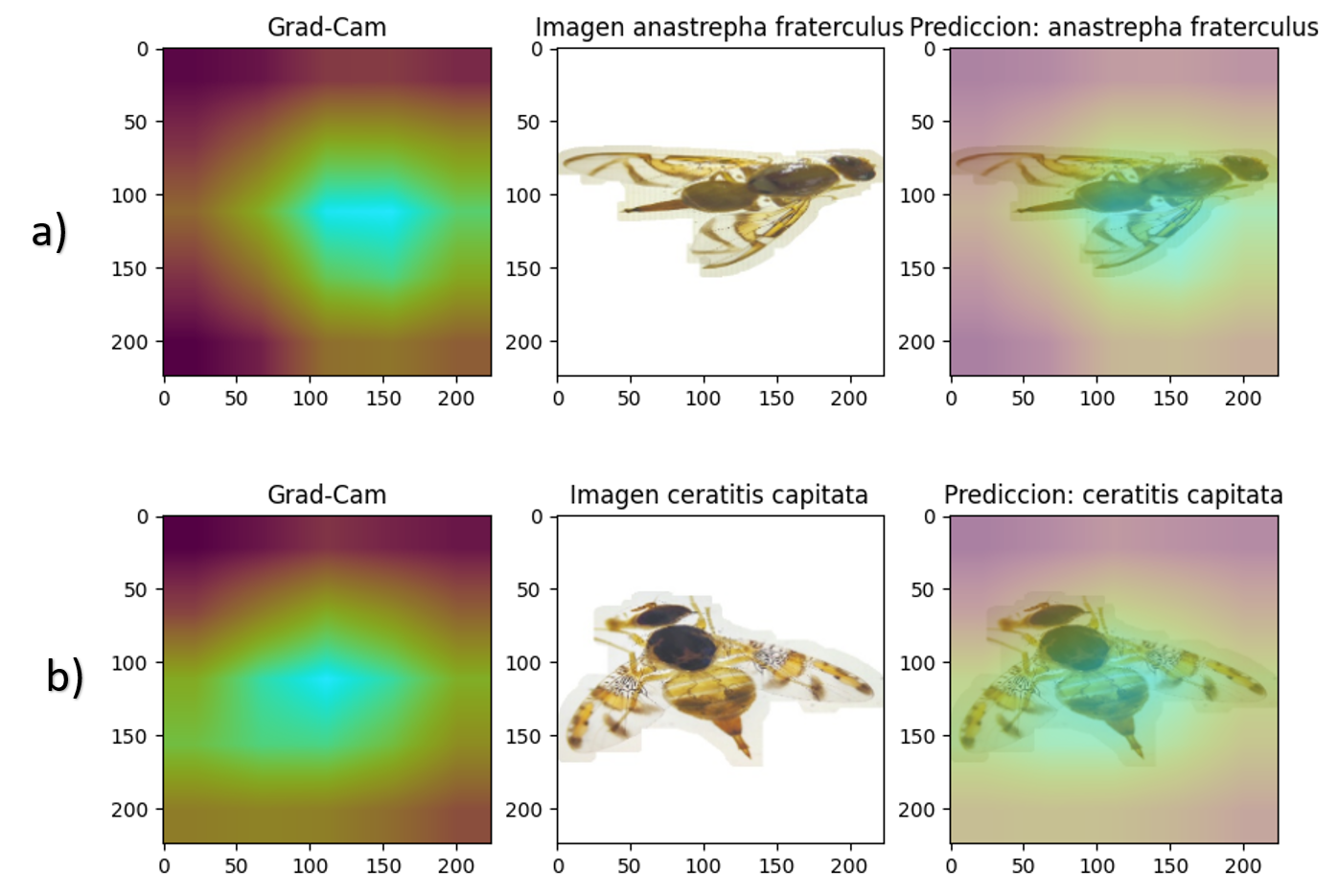}
    \caption{Grad-Cam of the Inception-V3 model. a) \textit{Anastrepha fraterculus}. b) \textit{Ceratitis capitata}.}
    \scriptsize \textit{Source:} Author's own creation
    \label{fig:caminception}
\end{figure}

Figures \ref{fig:camvgg16}, \ref{fig:camvgg19}, \ref{fig:caminception} display the areas of interest identified by each of our models for classifying the same images from Figure \ref{fig:moscaprueba}. These results allow us to analyze which visual features were key for the models to assign a class to each image.

The Grad-CAM method allows us to identify the areas of the images that the models consider most relevant for making their predictions. In this analysis, we used images in which all three models classified correctly, which allowed us to evaluate the activated regions in each case.

In Figures \ref{fig:camvgg16} and \ref{fig:camvgg19}, corresponding to the VGG16 and VGG19 models, we observe that both share similar patterns when selecting areas of interest. The activated regions are more widely distributed, focusing on multiple subareas within the main parts of the object, such as the wings and thorax. This suggests that both models tend to consider more diverse information when making their predictions, utilizing various features present in the image.

On the other hand, in Figure \ref{fig:caminception}, the Inception V3 model shows a different approach by focusing its attention on a single predominant region. This area encompasses essential features, such as the wings and thorax, which are determinant for classification. This behavior could explain the model's excellent performance when evaluated with the test data, achieving 93\% in the F1-score metric, as it effectively utilizes the most relevant features to make its decisions.

\subsection{Criteria for Model Selection.}
The criteria considered for model selection include classification time, performance metrics, areas of interest identified by the models, and their performance in uncontrolled scenarios. These criteria are based on the results obtained from evaluating the models with test images, consisting of 74 images per class under study.

\subsubsection{Evaluation of Classification Time.}
When comparing classification times, the Inception-V3 model proved to be the most efficient, with a total time of 15.04 seconds. It was followed by the VGG19 model, with a time of 40.83 seconds, while VGG16 recorded a time of 45.13 seconds. These results position Inception-V3 as the most efficient model in terms of classification speed.

\subsubsection{Evaluation of Performance Metrics.}
In terms of performance metrics, Inception-V3 excelled with an F1-score of 93\%, reflecting its superior ability to learn and generalize the patterns associated with the studied species. In contrast, VGG16 and VGG19 achieved F1-scores of 0.82\%. These findings suggest that Inception-V3 offers higher precision and robustness in classification.

\subsubsection{Evaluation Using Grad-CAM.}
In Section \ref{sec:cam}, the graphical representations of the image regions used by the models to make their predictions are presented, using the Grad-CAM technique. This technique highlights the areas of the image that are relevant for each prediction and is shown in the third column of Figures \ref{fig:camvgg16}, \ref{fig:camvgg19}, and \ref{fig:caminception}, where Grad-CAM is overlaid on the original image, facilitating the visualization of the areas on which each model focuses its attention.

Upon analyzing the results, it is observed that Inception-V3 covers a greater amount of information, including regions with key morphological characteristics. This demonstrates that Inception-V3 is the model that best utilizes these features to make its classifications.

The goal of this work is to develop a model capable of replicating the criteria used by experts in species classification. The results obtained confirm that Inception-V3 is the most suitable model for identifying relevant morphological areas, validating its suitability for this task.

\subsubsection{Evaluation in Uncontrolled Scenarios.}

For the evaluation of uncontrolled scenarios, images were selected from the internet with the goal of more thoroughly assessing the robustness and generalization capability of our models. We used a set of images manually collected from various online sources. These images represent real-world scenarios, including flies perched on fruits or fabrics, as well as images with uniform single-color backgrounds.

The final dataset consisted of 36 images, divided equally into 18 images for each class. This test allowed us to analyze how the models managed to generalize the specific features of the species under study when faced with data that were not part of the initial training. The results obtained are crucial for evaluating the performance of the models in uncontrolled contexts.

The results of the models with this dataset are shown in Tables \ref{tab:ivgg16}, \ref{tab:ivgg19}, and \ref{tab:iv3}.
\begin{table}[htbp]  
	\centering  
	\caption{Results with internet data from the VGG16 model.}  
	\label{tab:ivgg16}  
	\begin{tabular}{|l|c|c|c|}  
		\hline  
		& \textbf{Precision} & \textbf{Recall} & \textbf{F1-score} \\  
		\hline  
		\textbf{\textit{A. fraterculus}} & 0.4 & 0.11 & 0.17 \\  
		\hline  
		\textbf{\textit{C. capitata}} & 0.48 & 0.83 & 0.61 \\  
		\hline  
		\textbf{Macro average} & 0.44 & 0.47 & \textbf{0.39} \\  
		\hline  
	\end{tabular}  
\end{table}
\begin{table}[htbp]  
	\centering  
	\caption{Results with internet data from the VGG19 model.}  
	\label{tab:ivgg19}  
	\begin{tabular}{|l|c|c|c|}  
		\hline  
		& \textbf{Precision} & \textbf{Recall} & \textbf{F1-score} \\  
		\hline  
		\textbf{\textit{A. fraterculus}} & - & 0 & - \\  
		\hline  
		\textbf{\textit{C. capitata}} & 0.5 & 1.00 & 0.67 \\  
		\hline  
		\textbf{Macro average} & - & 0.5 & - \\  
		\hline  
	\end{tabular}  
\end{table}
\begin{table}[htbp]  
	\centering  
	\caption{Results with internet data from the Inception-V3 model.}  
	\label{tab:iv3}  
	\begin{tabular}{|l|c|c|c|}  
		\hline  
		& \textbf{Precision} & \textbf{Recall} & \textbf{F1-score} \\  
		\hline  
		\textbf{\textit{A. fraterculus}} & 0.73 & 0.61 & 0.67 \\  
		\hline  
		\textbf{\textit{C. capitata}} & 0.67 & 0.78 & 0.72 \\  
		\hline  
		\textbf{Macro average} & 0.7 & 0.69 & \textbf{0.69} \\  
		\hline  
	\end{tabular}  
\end{table}

This study aims to develop a robust model that demonstrates its effectiveness in species classification. The results obtained highlight Inception-V3 as the model with the highest robustness and best classification time, successfully performing the task of fruit fly species classification.

%% file: contenido/conclusiones/main_conclusiones.tex
\section{Conclusions.}  
In this study, various classification models were evaluated with the goal of classifying fruit fly species in segmented images. Among the models tested, Inception-V3 demonstrated the highest efficiency, achieving a total classification time of 15.04 seconds for 148 samples and an F1-score of 93\%, indicating its suitability for species classification in terms of both accuracy and speed.

Although Inception-V3 showed strong performance in controlled conditions and with poorly segmented images, its generalization capability on a set of randomly selected images from the internet was limited, yielding an F1-score of 69\%. This result emphasizes the importance of incorporating more diverse conditions in future studies to enhance the model's applicability in real-world environments.

The visual analysis conducted with Grad-Cam revealed that Inception-V3 was able to identify key morphological features associated with the species, further supporting its ability to extract relevant characteristics for classification. This, combined with its performance on well-segmented data and additional tests, justifies its selection as the most appropriate model for the task.

In conclusion, Inception-V3 emerges as a promising model for species classification due to its balance of accuracy and speed, along with its partial adaptability to diverse scenarios. However, the findings also highlight the necessity of continuing to explore strategies to improve the model's generalization in uncontrolled conditions, expanding its potential for real-world applications.

%% file: contenido/agradecimientos/agradecimientos.tex
\section*{Acknowledgments.}  
We express our sincere gratitude to ``El Servicio Nacional de Sanidad Agraria del Perú'' for providing us with access to their laboratory, which was essential for capturing the images used in constructing the dataset, as well as for the technical support provided in the classification of the species. We also extend our gratitude to ``Laboratorio de Algoritmos y Análisis de Datos''  at the ``Universidad Nacional de San Antonio Abad del Cusco'' for their valuable guidance and support during the drafting of this article.